\documentclass[12pt]{article}

\usepackage{graphicx}
\usepackage{amsmath}
\usepackage{url}
\usepackage{hyperref}

\begin{document}
\title{LLMs and the Madness of Crowds}
\author{William F. Bradley\thanks{Mirabolic Consulting}}
\maketitle

\begin{abstract}
We investigate the patterns of incorrect answers produced by large language models (LLMs) during evaluation. These errors exhibit highly non-intuitive behaviors unique to each model. By analyzing these patterns, we measure the similarities between LLMs and construct a taxonomy that categorizes them based on their error correlations. Our findings reveal that the incorrect responses are not randomly distributed but systematically correlated across models, providing new insights into the underlying structures and relationships among LLMs.
\end{abstract}

\section{Introduction}

(With apologies to Charles Mackay~\cite{mackay2012extraordinary})

The performance of large language models (LLMs) is frequently evaluated using multiple-choice tests. When an LLM's inference is performed with a positive temperature, posing the same problem repeatedly will yield a distribution across the possible answers. If the LLM performs well, we would expect most of the probability mass to lie on the correct answer; if it performs poorly, we might expect the distribution to be more uniform across all the answers.

However, this intuition does not always hold. To better understand the actual behavior of LLMs, we provide several detailed examples in Section~\ref{sec:examples}. In Section~\ref{sec:madness}, we perform a more comprehensive analysis at scale and use the results to suggest a taxonomy of LLMs.

\section{Non-Uniformity in Answer Distributions} \label{sec:examples}

In this section, we provide several detailed examples LLM behavior in multiple-choice tests. We postpone a more substantive statistical analysis for Section~\ref{sec:madness}.

We selected nine problems at random from the NeoSQuAD dataset~\cite{bradley2024enhancing}. NeoSQuAD is a multiple choice LLM evaluation where each problem contains a context paragraph, a question, and three answers. We suppressed the context and prompted OpenAI's \texttt{gpt-4o-2024-08-06} LLM with the question and the answers in a random order. We repeated this process 1,200 times per problem to produce a histogram across answers for each problem.

\begin{figure}[htbp]
  \centering
  \includegraphics[width=0.8\textwidth]{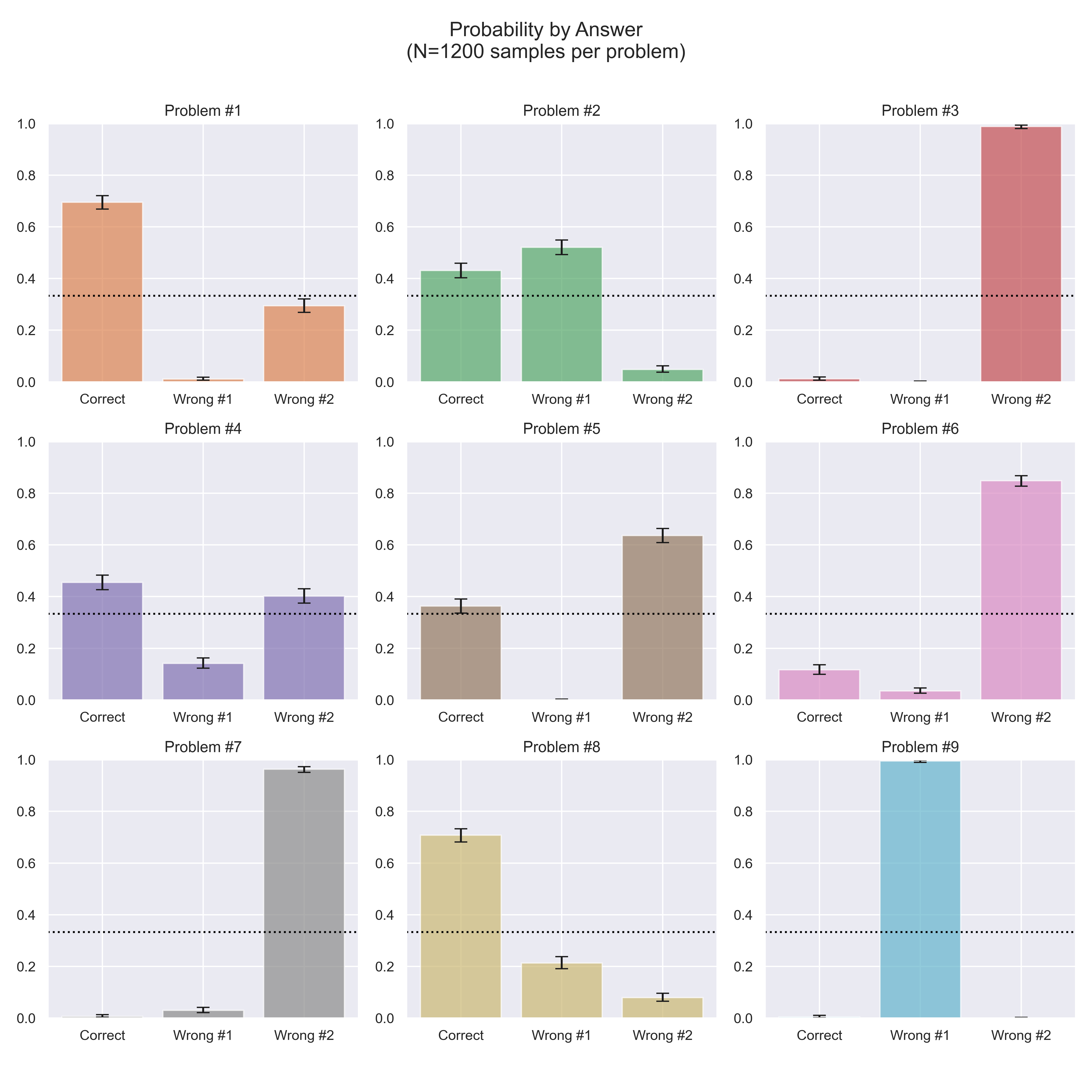}
  \caption{Histograms of answers selected by \texttt{gpt-4o-2024-08-06} for nine multiple-choice problems without context.}
  \label{fig:histo_by_answer}
\end{figure}

As shown in Figure~\ref{fig:histo_by_answer}, the histograms are highly non-uniform.\footnote{One can also use this data to measure the effect of the answer's location on the distribution (e.g., how often does the LLM select ``A''?). The distribution varied by question, but was substantially more uniform than the distribution by answer.} In several cases (Problems 3, 7, and 9), the LLM chose the same (incorrect) answer over 99\% of the time. Similar effects are sometimes attributed to data leakage into the LLM's pretraining data, but the fact that the LLM prefers the wrong answer argues against that explanation.

\begin{figure}[htbp]
  \centering
  \includegraphics[width=0.8\textwidth]{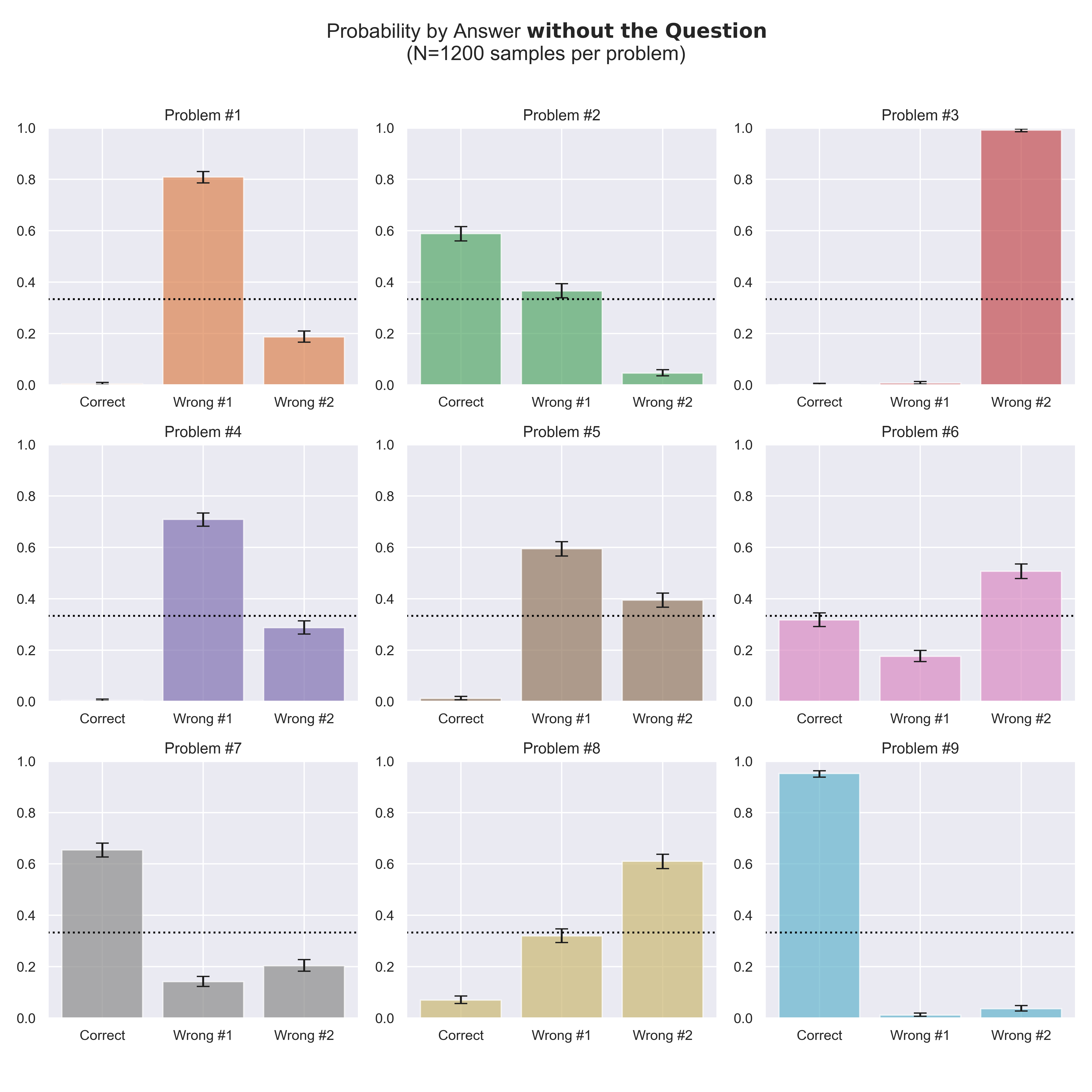}
  \caption{Histograms of answers selected by \texttt{gpt-4o-2024-08-06} for nine multiple-choice problems without questions or context. Labels ``correct'' and ``incorrect'' are mirrored from Figure~\ref{fig:histo_by_answer} for comparability, although they are not strictly meaningful without a question.}
  \label{fig:histo_by_answer_no_question}
\end{figure}

We can even remove the question and simply ask the LLM to choose an answer from the list. While it is no longer meaningful to refer to a ``correct'' or ``incorrect'' answer in this context, the non-uniformity of the histograms remains striking, as shown in Figure~\ref{fig:histo_by_answer_no_question}. Sometimes the LLM remains confident in the same answer (Problem 3), and sometimes it switches its preference to a different answer (Problem 9).

\begin{figure}[htbp]
  \centering
  \includegraphics[width=0.8\textwidth]{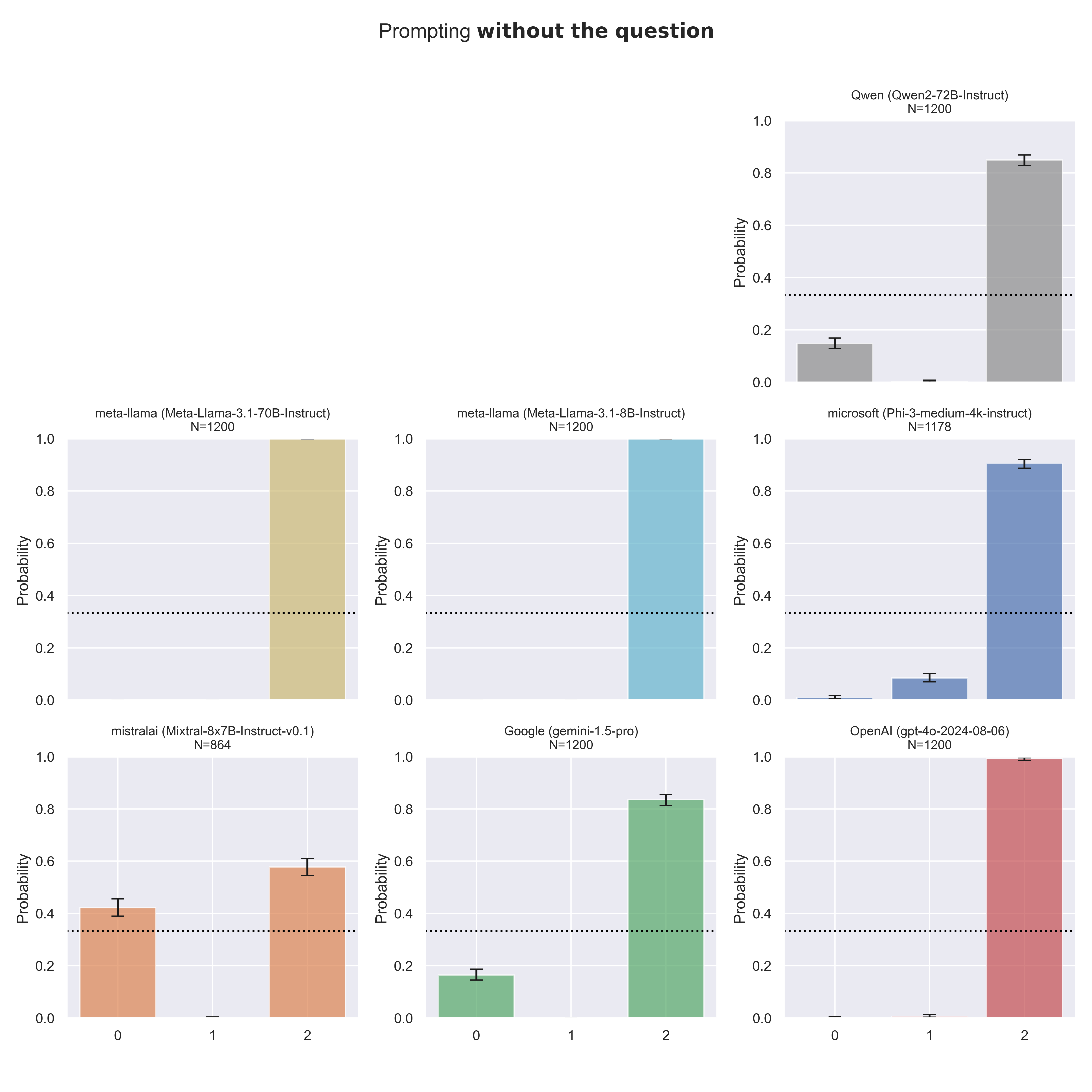}
  \caption{Histograms of answers selected for Question~3 by seven different LLMs.}
  \label{fig:histo_by_model_no_question}
\end{figure}

Given that LLMs preferentially select particular answers, do different LLMs tend to pick the \emph{same} favorites? Because our first LLM responded so confidently (and incorrectly) to Question~3, we examined its behavior across different models, as shown in Figure~\ref{fig:histo_by_model_no_question}.

A classic and powerful technique in machine learning is building an ``ensemble'' of multiple models~\cite{mienye2022ensemble}. One implication of this correlated non-uniformity is that ensembling LLMs may prove much less effective for LLMs than it does with other models. (Note that ensembling between LLMs is different from a mixture-of-experts within a single LLM~\cite{cai2024survey}.)

\section{LLM Correlations and Taxonomy} \label{sec:madness}

Given the histograms in Section~\ref{sec:examples}, an LLM's incorrect answers may prove more interesting than the correct ones. Moreover, Figure~\ref{fig:histo_by_model_no_question} suggests that these incorrect answers between different LLMs may be correlated. We use the MMLU-Pro evaluation to investigate this behavior more carefully~\cite{wang2024mmlu}.

MMLU-Pro is a multiple-choice test with over 12,000 questions across various subjects. It consists of a difficult core selected from the larger MMLU test~\cite{hendrycks2020measuring}. The number of options per question varies, but many problems offer ten options. The TIGER AI Lab at the University of Waterloo designed the MMLU-Pro and, as of October 2024, has recorded the problem-level performance of 37 LLMs on the test, with publicly available data~\cite{tiger2024mmlupro}.

\begin{figure}[htbp]
  \centering
  \includegraphics[width=1.2\textwidth]{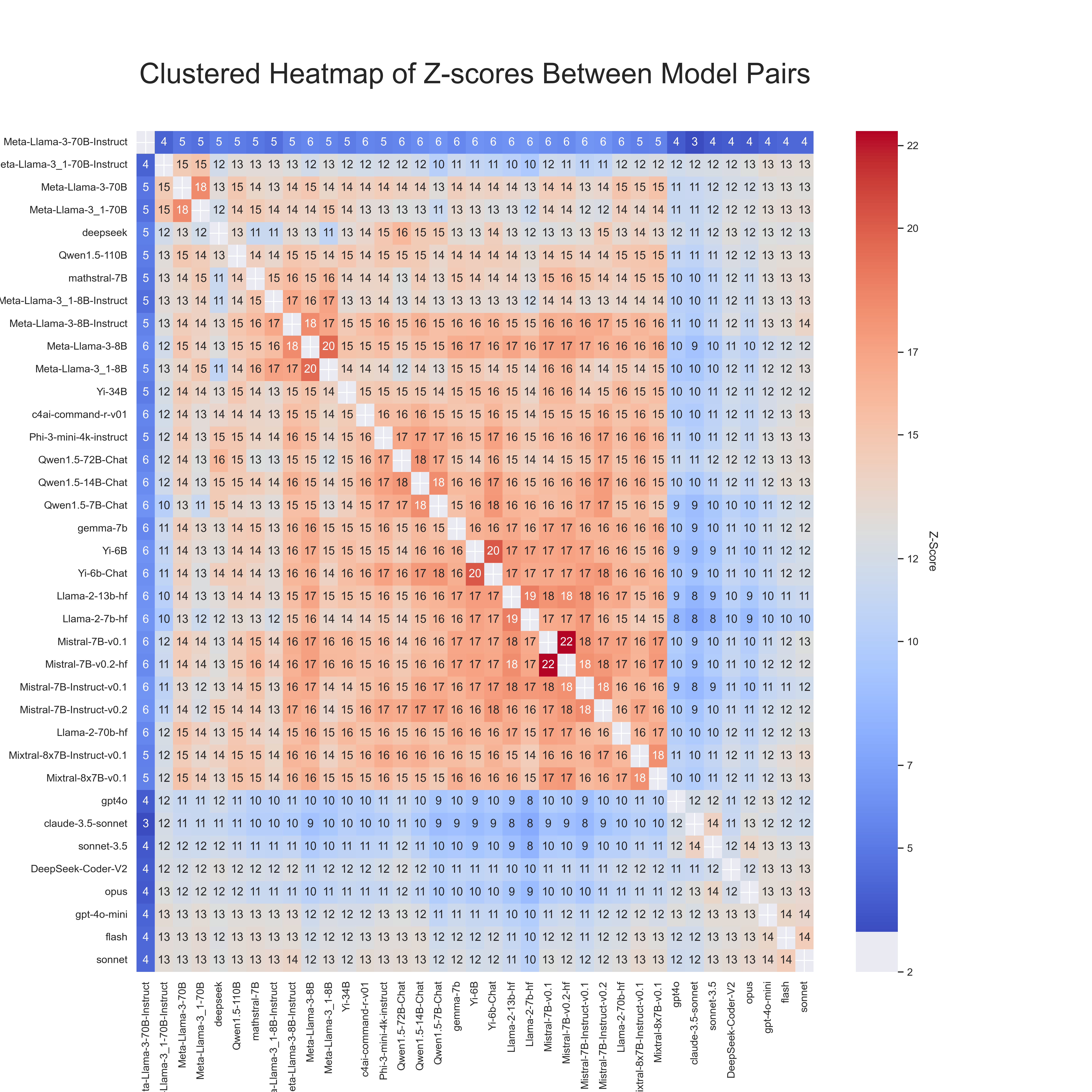}
  \caption{LLM similarity measured by correlation of errors on the MMLU-Pro test. Correlation is measured by $z$-score.}
  \label{fig:madness}
\end{figure}

We use these responses to measure the similarity between LLMs. Our approach is as follows: Given two LLMs evaluated on MMLU-Pro, we restrict our analysis to the set of problems where neither LLM selects the correct answer. For a particular problem with $k$ options, if neither LLM chooses the correct option and they select answers independently and uniformly from the remaining options, there is a $1/(k-1)$ chance of a match. (Note that $k$ varies between problems in MMLU-Pro.) We compute the mean and variance of this distribution, allowing us to calculate a $z$-score. This statistic essentially measures ``How unlikely is our observed correlation, measured in standard deviations above the mean?'' We compute the $z$-score between all pairs of LLMs and show the results in Figure~\ref{fig:madness}.

The most striking feature of Figure~\ref{fig:madness} is the scale of the $z$-scores. The smallest $z$-score we observe is 2.97, and the median is 13.15. In other words, in terms of selecting the same incorrect answers, all 37 LLMs behave similarly, and most are strikingly similar.

Note that there is no a priori guarantee that two LLMs will make mistakes on the same problems, which could limit data for comparison. Fortunately, we have sufficient data: Across the 37 LLMs considered, each pair had at least 994 common errors, with a median of 4,592.5.

\begin{figure}[htbp]
  \centering
  \includegraphics[width=1.0\textwidth]{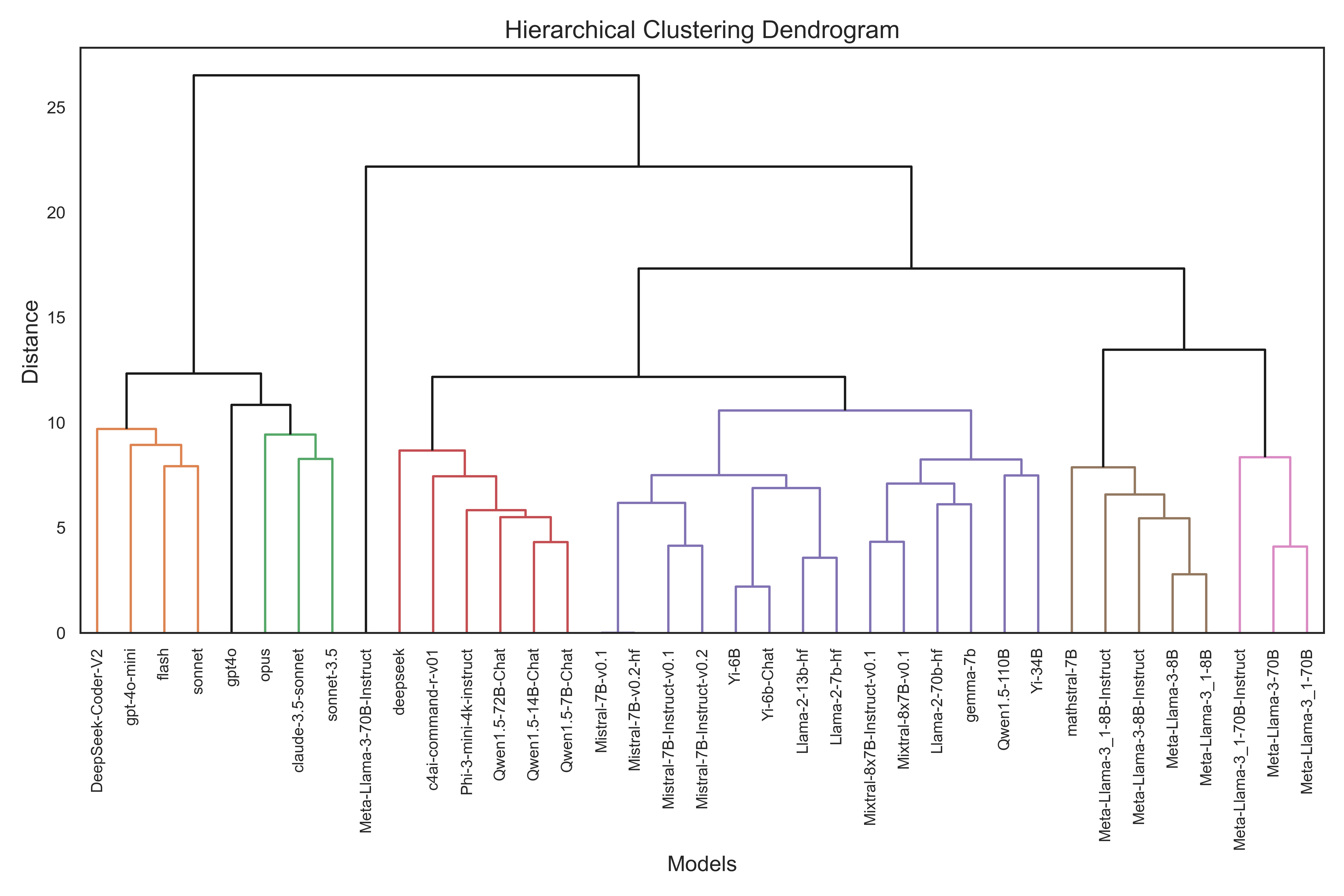}
  \caption{Hierarchical clustering of LLMs based on correlated errors.}
  \label{fig:dendrogram}
\end{figure}

These pairwise correlations between LLMs can be combined to illuminate the broader structure between families of models. As an initial exploration of this structure, we performed hierarchical clustering, as shown in Figure~\ref{fig:dendrogram}, with the LLMs ordered accordingly.

Examining the dendrogram, we can recognize several interpretable phenomena. Broadly speaking, there is a major split between proprietary models (the orange and green clusters on the left) and open-source models, with Meta's Llama models behaving like other open-source models. Additionally, the green cluster corresponds to Anthropic's models; the brown cluster corresponds to the smaller Llama models (and \texttt{mathstral-7B}); the pink cluster corresponds to the larger Llama models. The pink and brown super-cluster tightly maps to the Llama models in general.

Two particular models warrant comment. First, \texttt{DeepSeek-Coder-V2} is unusual, perhaps because it is fine-tuned on computer code rather than human text. This effect is clearer in Figure~\ref{fig:madness} than in Figure~\ref{fig:dendrogram}. (Note that this data is collected from the 236B parameter version of \texttt{DeepSeek-Coder-V2}.) Second, \texttt{Meta-Llama-3-70B-Instruct} is significantly different from all other models. This dissimilarity is striking, as neither \texttt{Meta-Llama-3\_1-70B-Instruct} nor \texttt{Meta-Llama-3-70B} demonstrate this effect. It follows that if we wish to choose a relatively ``independent'' response, \texttt{Meta-Llama-3-70B-Instruct} may be the best choice for \emph{any} other LLM.

We also mention that performing hierarchical clustering involves numerous decisions with complex trade-offs. For instance, we used Ward's~\cite{ward1963hierarchical} algorithm for linkage in Figure~\ref{fig:dendrogram}, as we felt it produced the most informative clusters. The UPGMA method~\cite{michener1957quantitative}, on the other hand, produces less distinct clusters but reveals that there is much greater diversity between proprietary models than between open-source models. (We hypothesize this effect may be caused by greater diversity of training data in the proprietary models.) In light of these different possibilities, we find it more helpful to treat the dendrogram's conclusions qualitatively rather than quantitatively.

\section{Universal Errors}

We briefly examine ``universal errors,'' i.e., problems that stump all or nearly all the LLMs.

First, if all 37 LLMs get the same question wrong, do they all get it wrong \emph{in the same way}? We have a partial affirmative answer in the previous section, but we were comparing LLMs pairwise. Of the 12,000+ problems in MMLU-Pro, there are 160 problems that all 37 LLMs answer incorrectly. We can ask what fraction of LLMs select the most popular answer.

To interpret that fraction, we begin by determining what the expected fraction would be if the LLMs selected answers uniformly at random. Given $N$ possible choices and $M$ selections, the expected maximum fraction of agreements on an answer is approximately $1/M + \sqrt{2\log(M)/NM}$ (per, e.g.,~\cite{mitzenmacher1995probability}). These questions all had 10 possible answers, hence $M=9$ wrong answers and $N=37$ models, so the expected maximal fraction of agreement is approximately $0.226$.

\begin{figure}[htbp]
  \centering
  \includegraphics[width=0.8\textwidth]{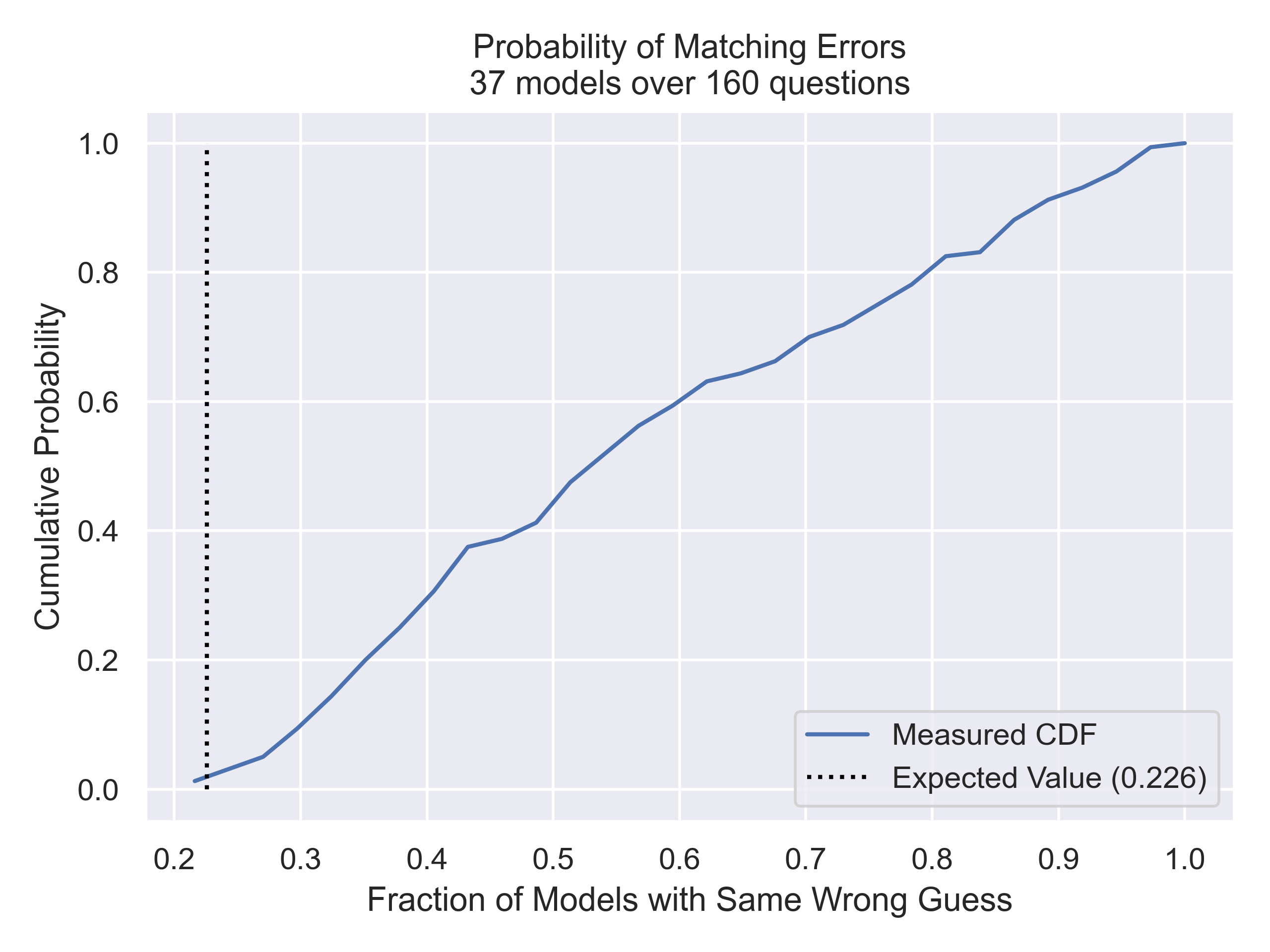}
  \caption{If every LLM is incorrect, what fraction of them agree on the same answer?}
  \label{fig:balls_and_bins}
\end{figure}

We measure the empirical fraction and display the cumulative distribution in Figure~\ref{fig:balls_and_bins}. The fraction appears to be uniformly distributed between 0.226 and 1; in other words, there is evident positive correlation, but LLMs still occasionally choose other answers.

Sometimes, however, all the LLMs \emph{do} choose the same wrong answer. There was such a question, along with six more questions where 36 of 37 LLMs agreed on the same wrong answer. In all but one case, either the ``correct'' answer was in error or the options included more than one correct answer~\cite{MMLUPro_Issue16_2024}. In other words, the LLMs agreed with each other because the test was flawed, which provides a different perspective on our interpretation. Nevertheless, this situation is exceptional; in most cases, the labeled answer is correct and unique.

\bibliographystyle{plain}
\bibliography{references}

\end{document}